\documentclass[journal]{IEEEtran}
\IEEEoverridecommandlockouts
\usepackage{color}
\usepackage{tikz}
\usetikzlibrary{shapes,arrows,arrows.meta}
\usepackage{mathtools}
\usepackage{amssymb}
\usepackage{lipsum}
\usepackage{float}
\usepackage{makecell}
\usepackage{longtable}
\usepackage{stfloats}
\usepackage{multirow}
\usepackage{amsmath}
\usepackage{bm}
\usepackage{algorithm}
\usepackage{algpseudocode}
\usepackage{booktabs}
\usepackage[numbers,sort&compress]{natbib}
\usepackage{balance}
\usepackage{soul}
\algdef{SE}[SUBALG]{Indent}{EndIndent}{}{\algorithmicend\ }%
\algtext*{Indent}
\algtext*{EndIndent}
\usepackage{xcolor}
\usepackage{threeparttable}
\usepackage{natbib}
\usepackage[breaklinks,citecolor=red,urlcolor=red,bookmarks=false,hypertexnames=true]{hyperref}
\allowdisplaybreaks
\usepackage{graphicx}
\usepackage{caption}
\usepackage{subfigure}
\begin{document}

\title{LXL: LiDAR Excluded Lean 3D Object Detection with 4D Imaging Radar and Camera Fusion}

\author{\IEEEauthorblockA{
Weiyi Xiong\IEEEauthorrefmark{1},
Jianan Liu\IEEEauthorrefmark{1},
Tao Huang,~\IEEEmembership{Senior Member,~IEEE,}\\
Qing-Long Han,~\IEEEmembership{Fellow,~IEEE,} 
Yuxuan Xia, and
Bing Zhu\IEEEauthorrefmark{2},~\IEEEmembership{Member,~IEEE}
}
\vspace{-5 mm}

\thanks{W.~Xiong and B.~Zhu are with the School of Automation Science and Electrical Engineering, Beihang University, Beijing 100191, P.R.~China. Email:
weiyixiong@buaa.edu.cn (W. Xiong);
zhubing@buaa.edu.cn (B. Zhu).}
\thanks{J.~Liu is with Vitalent Consulting, Gothenburg, Sweden. Email: jianan.liu@vitalent.se.}
\thanks{T.~Huang is with the College of Science and Engineering, James Cook University, Cairns QLD 4878, Australia. Email: tao.huang1@jcu.edu.au.}
\thanks{Q.-L.~Han is with the School of Science, Computing and Engineering Technologies, Swinburne University of Technology, Melbourne, VIC 3122, Australia. Email: qhan@swin.edu.au.}
\thanks{Y.~Xia is with the Department of Electrical Engineering, Chalmers University of Technology, Gothenburg, Sweden. Email: yuxuan.xia@chalmers.se.}
\thanks{\IEEEauthorrefmark{1}Both authors contribute equally to the work and are co-first authors.}
\thanks{\IEEEauthorrefmark{2}Corresponding author.}
\thanks{This paper has been accepted by IEEE Transactions on Intelligent Vehicles. Digital Object Identifier 10.1109/TIV.2023.3321240}
}

\markboth{IEEE Transactions on Intelligent Vehicles}
{\MakeLowercase{\textit{et al.}}: Demo of IEEEtran.cls for IEEE Journals}

\maketitle

\begin{abstract}
As an emerging technology and a relatively affordable device, the 4D imaging radar has already been confirmed effective in performing 3D object detection in autonomous driving. Nevertheless, the sparsity and noisiness of 4D radar point clouds hinder further performance improvement, and in-depth studies about its fusion with other modalities are lacking.
On the other hand, as a new image view transformation strategy, ``sampling" has been applied in a few image-based detectors and shown to outperform the widely applied ``depth-based splatting" proposed in Lift-Splat-Shoot (LSS), even without image depth prediction.
However, the potential of ``sampling" is not fully unleashed. 
This paper investigates the ``sampling" view transformation strategy on 
the camera and 4D imaging radar fusion-based 3D object detection. 
LiDAR Excluded Lean (LXL) model, predicted image depth distribution maps and radar 3D occupancy grids are generated from image perspective view (PV) features and radar bird's eye view (BEV) features, respectively. They are sent to the core of LXL, called ``radar occupancy-assisted depth-based sampling", to aid image view transformation.
We demonstrated that more accurate view transformation can be performed by introducing image depths and radar information to enhance the “sampling” strategy.
Experiments on VoD and TJ4DRadSet datasets show that the proposed method outperforms the state-of-the-art 3D object detection methods by a significant margin without bells and whistles. Ablation studies demonstrate that our method performs the best among different enhancement settings.

\end{abstract}

\begin{IEEEkeywords}
4D imaging radar, camera, multi-modal fusion, 3D object detection, deep learning, autonomous driving
\end{IEEEkeywords}

\IEEEpeerreviewmaketitle
\setcitestyle{square}

\section{Introduction}
\label{introduction}

Perception plays a pivotal role in autonomous driving, since its subsequent procedures (e.g., trajectory prediction, motion planning and control) rely %heavily 
on accurately perceiving the environment. Key tasks in this domain encompass segmentation \cite{RadarInsSeg}\cite{RadarInsSeg2}, object detection \cite{RaLiBEV}, and tracking \cite{GNN-PMB}\cite{Sun2023vision}, with 3D object detection being the most widely researched area. 

The approach to performing 3D object detection in autonomous driving varies based on the type of sensor employed. LiDARs, cameras, and radars are commonly utilized sensors characterized by distinct data structures and properties. LiDAR data is in the form of point clouds, providing precise 3D geometric information regarding an object's shape, size, and position. Meanwhile, camera images offer dense and regular data, supplying rich semantic information. 
However, the high cost of LiDARs prohibits their widespread adoption in household vehicles, and cameras are susceptible to challenging lighting and weather conditions \cite{3d_object_det_survey_1}. 

\begin{table}[]
\centering
\caption{The comparison of conventional automotive radars (3D radars) and 4D imaging radars (4D radars) \cite{4D_radar_overview}.}\label{radar comparison}
\begin{tabular}{l|ll}
\hline
Characteristics & 3D Radars & 4D Radars \\ \hline
Ability to Measure Elevation Angles & Weaker & Stronger \\
Resolution & Lower & Higher \\
Clutter / Noise & More & Less \\
Number of Points & Smaller & Larger \\ \hline
\end{tabular}
\end{table}

In contrast, radars are cost-effective and resilient to external factors, making them vital for robust detection in current advanced driver assistance systems (ADAS) and autonomous driving \cite{automotive_radar_survey}. Moreover, radars hold promise for future applications in cooperative perception \cite{cooperative_perception_survey_tits}. 
However, as shown in Table \ref{radar comparison}, conventional automotive radars, when used alone, lack height information and generate sparse point clouds, posing challenges for 3D object detection. The emergence of 4D imaging radars has led to the generation of higher-resolution 3D point clouds \cite{4D_radar_overview}. Although there is still a notable disparity in density and quality compared to LiDAR point clouds, several studies \cite{VoD}\cite{RPFA-Net}\cite{RadarMFNet} have explored 4D radar-based detection and demonstrated its feasibility.

In 3D object detection, researchers increasingly turn to multi-modal fusion techniques to overcome the limitations associated with single-modal data to improve the overall performance. One prominent approach involves independently extracting bird's-eye-view (BEV) features from different sensor modalities and integrating them into a unified feature map.
The utilization of BEV representation offers numerous advantages. Firstly, it enables more efficient processing compared to point-based or voxel-based methods. Additionally, leveraging mature 2D detection techniques can facilitate learning processes. Furthermore, occlusion, a common challenge in other representations like the range-view, is mitigated in BEV \cite{3d_object_det_survey_2}. Notably, using BEV representation simplifies and enhances the effectiveness of multi-modal fusion strategies \cite{3d_object_det_survey_3}.

Despite the benefits of using BEV representation for multi-modal fusion in 3D object detection, transforming images from a perspective view (PV) to BEV is intricate. Current approaches can be categorized into geometry-based \cite{CaDDN}\cite{BEVDet}\cite{M2BEV} and network-based methods \cite{BEVFormer}\cite{PolarFormer}. Geometry-based approaches, relying on explicit utilization of calibration matrices, offer a more straightforward learning process than network-based approaches.  One widely employed geometry-based method is ``depth-based splatting". Initially introduced in Lift-Splat-Shoot (LSS) \cite{LSS}, this method lifts image pixels into 3D space guided by predicted pixel depth distributions.
Several enhancements have been proposed to improve its performance. For example, BEVDepth \cite{BEVDepth} generates a ``ground-truth" depth map from LiDAR points to supervise image depth prediction, whereas CRN \cite{CRN} employs a 2D radar occupancy map to assist view transformation. Another approach, called ``sampling", has demonstrated superior performance even without explicit depth prediction, as exemplified by Simple-BEV \cite{Simple-BEV}. 
However, unlike the common practice of ``splatting", few studies explored the combination of ``sampling" with predicted depths. Moreover, the potential of ``sampling" in conjunction with other modalities remains largely unexplored, indicating untapped opportunities for further improvement in this area.

Despite growing interest in multi-modal fusion techniques for 3D object detection, the specific integration of 4D imaging radar and cameras has received limited attention in the existing literature.
Existing methods designed for LiDAR-based fusion, such as the popular ``splatting" approach, is applicable to 4D imaging radar and camera fusion, but their enhancements like BEV-Depth \cite{BEVDepth} may fail due to the distinct characteristics of radar point clouds. Specifically, when point clouds from 4D radars instead of LiDARs can be accessed, the generated depth maps in BEV-Depth may suffer from sparsity and imprecision of radar points.
Additionally, methods devised specifically for radars, such as the technique presented in CRN \cite{CRN}, may introduce computational complexity and hinder the real-time inference capabilities of the model. 
Therefore, there is a clear need to address this research gap by developing novel fusion methods tailored for 4D imaging radar and camera fusion. 
In this study, we aim to enhance the existing ``sampling" method by leveraging the unique advantages of 4D imaging radar. 
By conducting extensive ablation studies, we show how 4D imaging radar can assist in image view transformation and demonstrate its impact on the overall 3D object detection performance. 
The contributions of this work are threefold:

\begin{itemize}
\item 
A LiDAR Excluded Lean (LXL) model is proposed
to perform 4D imaging radar and camera fusion-based 3D object detection.
This is an early attempt in this field, and serves as the latest benchmark for subsequent studies.

\item A ``radar occupancy-assisted depth-based sampling" feature lifting strategy is proposed in our view transformation module. It 
utilizes bi-linear sampling to get image features for pre-defined voxels, followed by two parallel operations: one combines image 3D features with the information from predicted image depth distribution maps, and the other exploits estimated radar 3D occupancy grids. 
This design enhances the underdeveloped ``sampling" strategy by introducing predicted depth distribution maps and radar 3D occupancy grids as assistance, leading to more precise feature lifting results.

\item Experiments show that LXL outperforms state-of-the-art models on the View-of-Delft (VoD) \cite{VoD} and TJ4DRadSet \cite{TJ4DRadSet} datasets by 6.7\% and 2.5\%, respectively, demonstrating the effectiveness of LXL.
In addition, comparisons of different feature lifting and radar assistance strategies are made through ablation studies, showing the superiority of the proposed view transformation module.

\end{itemize}

The rest of the paper is organized as follows. Section \ref{related work} reviews recent works on camera-based, camera and ordinary automotive radar fusion-based, and 4D imaging radar-based 3D object detection methods. Section \ref{method} details our proposed model, with focus on the view transformation module. Experimental settings and performances of our model and the corresponding analysis are provided in section \ref{experiment}. Finally, we summarize the work in this paper and point out the future research direction in Section \ref{conclusion}.

\section{Related Work}\label{related work}

\subsection{3D Object Detection with Cameras}
The camera-based 3D object detection work can be mainly categorized into three types. 

The first type involves directly estimating 3D bounding boxes based on image PV features \cite{M3D-RPN}. However, due to the inherent lack of depth information in images, the performances of these methods are limited.

The second type focuses on \textit{directly} transforming PV features into BEV and predicting bounding boxes on the top-down view \cite{BEVFormer}\cite{PolarFormer}. This approach requires additional information, e.g. pixel heights or depths, to achieve accurate view transformation. %A classical algorithm, 
Inverse perspective mapping (IPM) \cite{IPM} projects PV features onto the BEV with the assumption that all pixels lie on the ground. However, its performance is limited as the assumption is not always true.
Other methods \cite{BEVFormer}\cite{PolarFormer} utilize transformers to learn view transformation and reduce the impact of inaccurate depth estimation.

The third type involves lifting pixels into a point cloud or voxels \cite{CaDDN}\cite{BEVDet}\cite{M2BEV}\cite{BEVDepth}\cite{Pseudo-LiDAR} and applying networks designed for LiDAR-based 3D object detection. These approaches assume or estimate pixel depth or depth distributions to guide the 2D-to-3D projection. Pseudo-LiDAR \cite{Pseudo-LiDAR} is a pioneering work that transforms images into point clouds based on regressed depth. However, these networks cannot be trained end-to-end, limiting their performance. CaDDN \cite{CaDDN}, BEVDet \cite{BEVDet}, and BEVDepth \cite{BEVDepth} employ a technique where they discretize the depth space into bins, and treat the depth estimation as depth bin classification. Subsequently, image features are lifted into voxels based on the estimated depth distribution. In contrast, M$^2$BEV \cite{M2BEV} adopts a different strategy by assuming a uniform depth distribution instead of predicting depth probabilities. This approach mitigates the computational burden while allowing for effectively lifting image features into voxels.

It is important to note that most of the methods belonging to the third type also detect objects on the BEV. Projecting image features from PV to BEV alleviates the occlusion problem and facilitates multi-modal fusion. As a result, these types of methods have become more prevalent in recent years.

\subsection{3D Object Detection with Camera and Ordinary Automotive Radar Fusion}
Ordinary automotive radars cannot measure height information, posing a challenge for models to estimate 3D bounding boxes accurately from 2D radar points solely. Consequently, researchers have explored the fusion of 3D radar and camera data for improved 3D object detection, with earlier works including  \cite{Cam3DRadFusion1}\cite{Cam3DRadFusion2}\cite{CenterFusion}.

Recent advancements in this field have introduced novel approaches to fuse camera and radar data \cite{radar_camera_od_seg_survey}. For instance, Simple-BEV \cite{Simple-BEV} lifts image pixels to 3D voxels and concatenates them with radar BEV features before reducing the height dimension. To enhance the model's resilience against modal failure, CramNet \cite{CramNet} transforms the image foreground into a point cloud and utilizes it along with radar points for subsequent bounding box prediction. Given the relatively low reliability of image depth estimation, ray-constrained cross-attention is proposed in CramNet to refine the 3D location of pixels. Another work, RCBEV \cite{RCBEV}, employs a spatial-temporal encoder to extract features from accumulated radar sweeps and introduces a two-stage multi-modal fusion strategy.
In contrast to many previous approaches focusing on feature-level fusion, RADIANT \cite{RADIANT} adopts a result-level fusion strategy, merging depth predictions from radar and camera heads to achieve lower localization errors. 

Attention mechanisms and transformers have been leveraged to enhance camera and radar fusion performance. MVFusion \cite{MVFusion} and CRN \cite{CRN} utilize cross-attention to fuse camera and radar features. Additionally, they employ one modal as a guide to process the other modal. Specifically, MVFusion \cite{MVFusion} derives a Semantic Indicator from images to extract image-guided radar features, while CRN \cite{CRN} projects radar points onto images and generates occupancy maps within the view frustum, facilitating the depth-based PV to BEV transformation of image features. CRAFT \cite{CRAFT} primarily focuses on image detection and utilizes radar measurements to refine image proposals through the Spatio-Contextual Fusion Transformer. TransCAR \cite{TransCAR} incorporates a transformer decoder where vision-updated queries interact with radar features.

\subsection{3D Object Detection with 4D Imaging Radars}

With advancements in 4D imaging radars, which can generate 3D point clouds, there is a possibility of regressing 3D bounding boxes using radar modality alone. However, despite the availability of a few datasets providing 4D radar point clouds \cite{VoD}\cite{TJ4DRadSet}\cite{Astyx} and even 4D radar tensors \cite{RADIal}\cite{K-radar}, the research in this area remains limited. Most of the existing works in this field incorporate modules from LiDAR-based models, such as using the backbone of SECOND \cite{SECOND} or the detection head of CenterPoint \cite{CenterPoint}. Nevertheless, 3D point clouds generated by 4D imaging radars are typically sparser and noisier than LiDAR point clouds, often leading to lower model performance \cite{VoD}.

For example, \cite{VoD} applies PointPillars \cite{PointPillars} to perform 3D object detection using 4D imaging radars and achieves reasonable results. RPFA-Net \cite{RPFA-Net} modifies the pillarization operation of PointPillars \cite{PointPillars} by replacing the PointNet \cite{PointNet} with a self-attention mechanism to extract global features, thereby enhancing the model's orientation estimation capability. Using a spatial-temporal feature extractor on multi-frame radar point clouds and employing an anchor-based detection head, RadarMFNet \cite{RadarMFNet} achieves more accurate detection than single-frame methods. In the most recent work, SMURF \cite{SMURF}, a multi-representation fusion strategy is adopted where the PointPillars \cite{PointPillars} backbone and kernel density estimation (KDE) extract different radar features in parallel, enabling the acquisition of enhanced feature maps of radar points.

To leverage the data from other sensors such as LiDARs and cameras, researchers start to explore the fusion of these modalities with 4D imaging radars to achieve improved results. For instance, InterFusion \cite{InterFusion} and M$^2$-Fusion \cite{M2-Fusion} employ attention mechanisms \cite{self-attentions} to fuse pillarized LiDAR and radar features. \cite{Cam4DRadFusion} utilizes self-supervised model adaptation (SSMA) blocks \cite{SSMA} for a pixel-level fusion of image, radar BEV, and radar front view (FV) features. Similarly, RCFusion \cite{RCFusion} incorporates an interactive attention module to fuse radar and image BEV features. 

As 4D radar and camera fusion-based 3D object detection remains an area requiring further investigation, this paper aims to inspire subsequent researchers to explore this domain.

\begin{figure*}
    \centering
    \includegraphics[scale=0.56]{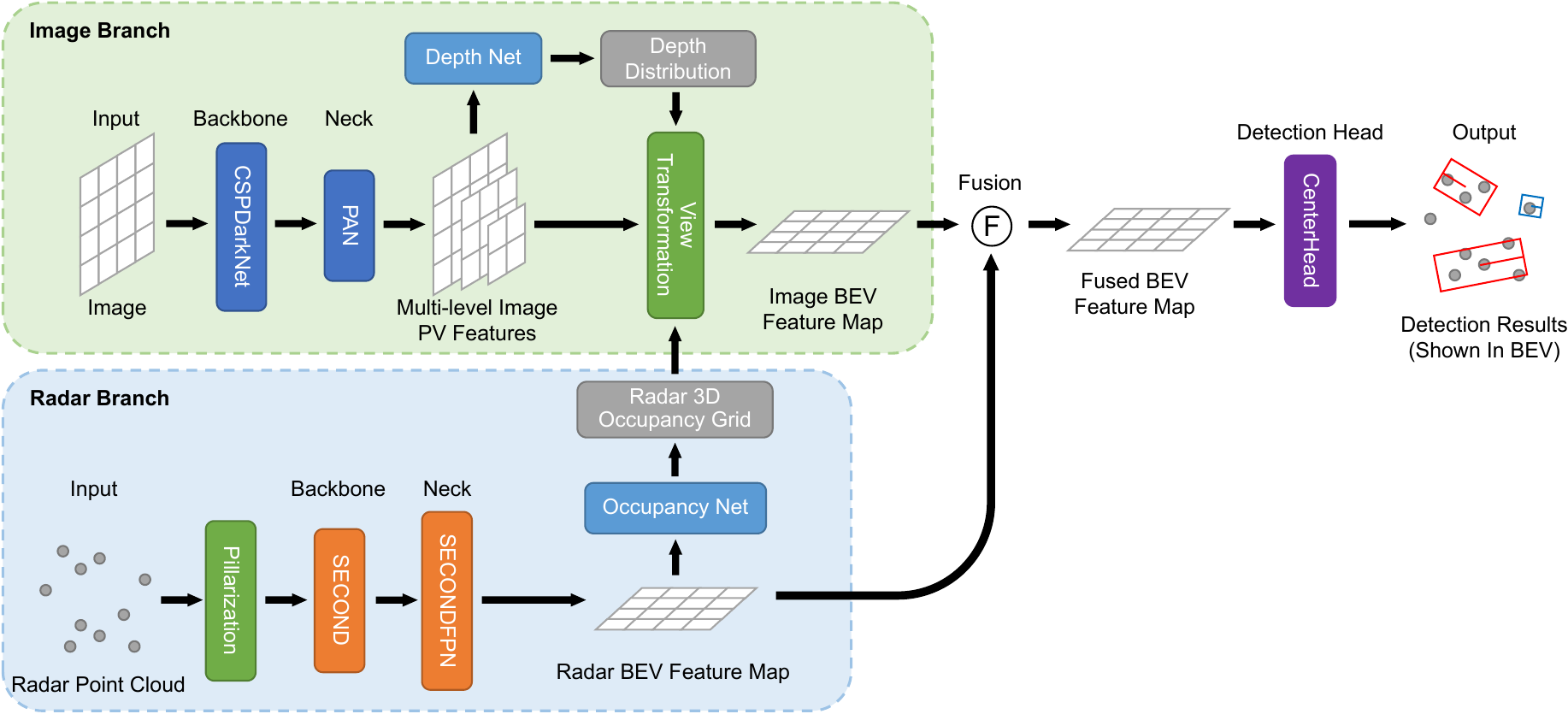}
    \caption{The overall architecture of LXL. The radar branch and image branch extract features of the corresponding modal, resulting in radar BEV feature maps and image PV feature maps. After that, radar 3D occupancy grids and image depth distribution maps are generated and sent to the view transformation module, assisting the image PV-to-BEV transformation. The resulted image BEV feature maps are fused with radar BEV feature maps, and bounding boxes are predicted from the output feature map of the fusion module.}
    \label{model}
\end{figure*}

\section{Proposed Method}        
\label{method}
\subsection{Overall Architecture}

The overall architecture of our model is depicted in Fig. \ref{model}. The model comprises four main components: the radar branch with an occupancy net, image branch with a depth net and a specially designed view transformation module, fusion module, and detection head. Each component plays a crucial role in the 3D object detection process:

\begin{itemize}
    \item[1.] The radar branch is responsible for processing radar point clouds as an input. It extracts radar BEV features, which capture essential information from the radar modality. Additionally, the radar branch generates 3D radar occupancy grids, representing the radar points' occupancy status within the scene.
    
    \item[2.] The image branch focuses on extracting multi-scale image PV features. These features encode relevant visual information from the image modality. We employ predicted image depth distribution maps and 3D radar occupancy grids to assist in transforming the image PV features into the BEV domain. Aligning the image features with the radar BEV representation enables effective fusion with the radar features.
    
    \item[3.] The fusion module is a key component in integrating the BEV features from both radar and image branches. It combines the complementary information each modality provides, allowing for enhanced object detection performance. The fusion process leverages the BEV features to generate a unified representation that captures the combined strengths of radar and image data.
    
    \item[4.] The detection head is responsible for bounding box regression and  classification for each potential object in the scene. It utilizes the fused features to estimate the 3D position, dimensions, orientation, and category of the objects. By leveraging the comprehensive information from both radar and image modalities, the detection head produces accurate predictions.
\end{itemize}

Further details regarding the proposed method are elaborated in the subsequent subsections.

\subsection{Radar Branch} \label{radar branch}
In the radar branch, the input radar point cloud is initially 
pillarized via the process employed in PointPillars \cite{PointPillars}.
Subsequently, the pillar representation is fed into the radar backbone and neck modules to extract relevant features. Following the widely referenced network SECOND \cite{SECOND}, the radar backbone and neck are constructed. The radar backbone extracts multi-level BEV features from the voxelized pillars. These features capture important spatial and contextual information inherent in the radar modality. The radar neck module then combines these multi-level features into a unified single-scale representation, facilitating subsequent fusion and analysis.
The obtained radar BEV feature maps serve two primary purposes within our model. Firstly, they are forwarded to the fusion module, where they are integrated with the image BEV features for effective object detection. Secondly, the radar BEV feature maps are utilized to predict radar 3D occupancy grids. The motivation behind the occupancy generation is discussed further in Section \ref{sec view trans}. 

To generate radar 3D occupancy grids, we employ an occupancy net. The radar BEV feature map, denoted as $\mathbf{F}_\text{BEV}^P$ with a shape $(X, Y, C_P)$, is fed into the occupancy net. Here, $X$ and $Y$ represent the dimensions of the feature map, and $C_P$ corresponds to the number of channels. In our framework, the height of the 3D occupancy grid, denoted as $Z$, is predefined. The occupancy net can be formulated as follows:
\begin{equation}
    \mathbf{O}_\text{3D}^P=\mathtt{Sigmoid}(\mathtt{Conv}_{C_P\rightarrow Z}(\mathbf{F}_\text{BEV}^P)),
\end{equation}
where $\mathbf{O}_\text{3D}^P\in \mathbb{R}^{X\times Y\times Z}$ is the predicted 3D occupancy grid and $\mathtt{Conv}_{a\rightarrow b}$ represents a $1\times1$ convolution layer with $a$ input channels and $b$ output channels. Note that during training, there are no direct supervise signals for occupancy prediction.

To transform the image view from PV to BEV, we leverage the radar 3D occupancy grids as an assistance. The specific details and the process of this transformation will be elaborated upon in Section \ref{sec view trans}.

\subsection{Image Branch} \label{image branch}

The image branch consists of several key modules: image backbone, neck, depth net, and view transformation module.

The image backbone extracts multi-level image PV features. The image neck is embraced to further enhance the features by mixing them at different scales. In our model, we employ the same architecture as YOLOX \cite{YOLOX} to achieve this design, utilizing CSPNet \cite{CSPNet} and PAN \cite{PAN}.

The depth net is implemented as a $1\times1$ convolutional layer for each multi-level image PV feature. Similar to many existing methods \cite{CaDDN}\cite{BEVDet}\cite{LSS}\cite{BEVDepth}, we discretize the depth space into multiple bins and treat the depth estimation task as a depth bin classification task. Consequently, the depth net outputs a depth probability distribution for each pixel. 
Given the image PV feature map of the $i$-th level, denoted as $\mathbf{F}_{i,\text{PV}}^I \in \mathbb{R}^{H_i\times W_i\times C_I}$, the depth distribution map $\mathbf{D}_i^I \in\mathbb{R}^{H_i\times W_i\times D}$ can be obtained as
\begin{equation}
    \mathbf{D}_i^I=\mathtt{Softmax}(\mathtt{Conv}_{C_I\rightarrow D}(\mathbf{F}_{i,\text{PV}}^I)),i=1,2,\cdots,N_\text{lvl},
\end{equation}
where $D$ represents the pre-defined number of depth bins, $N_\text{lvl}$ denotes the number of levels and $\mathtt{Softmax}(\cdot)$ is applied along the depth dimension.

The final module in the image branch is the view transformation module which is also the core of LXL. Its primary objective is to lift the image PV features into a 3D space and compress the height dimension. The detailed workings of this module will be elaborated upon in Section \ref{sec view trans}.

\begin{figure*}
    \centering
    \includegraphics[scale=0.56]{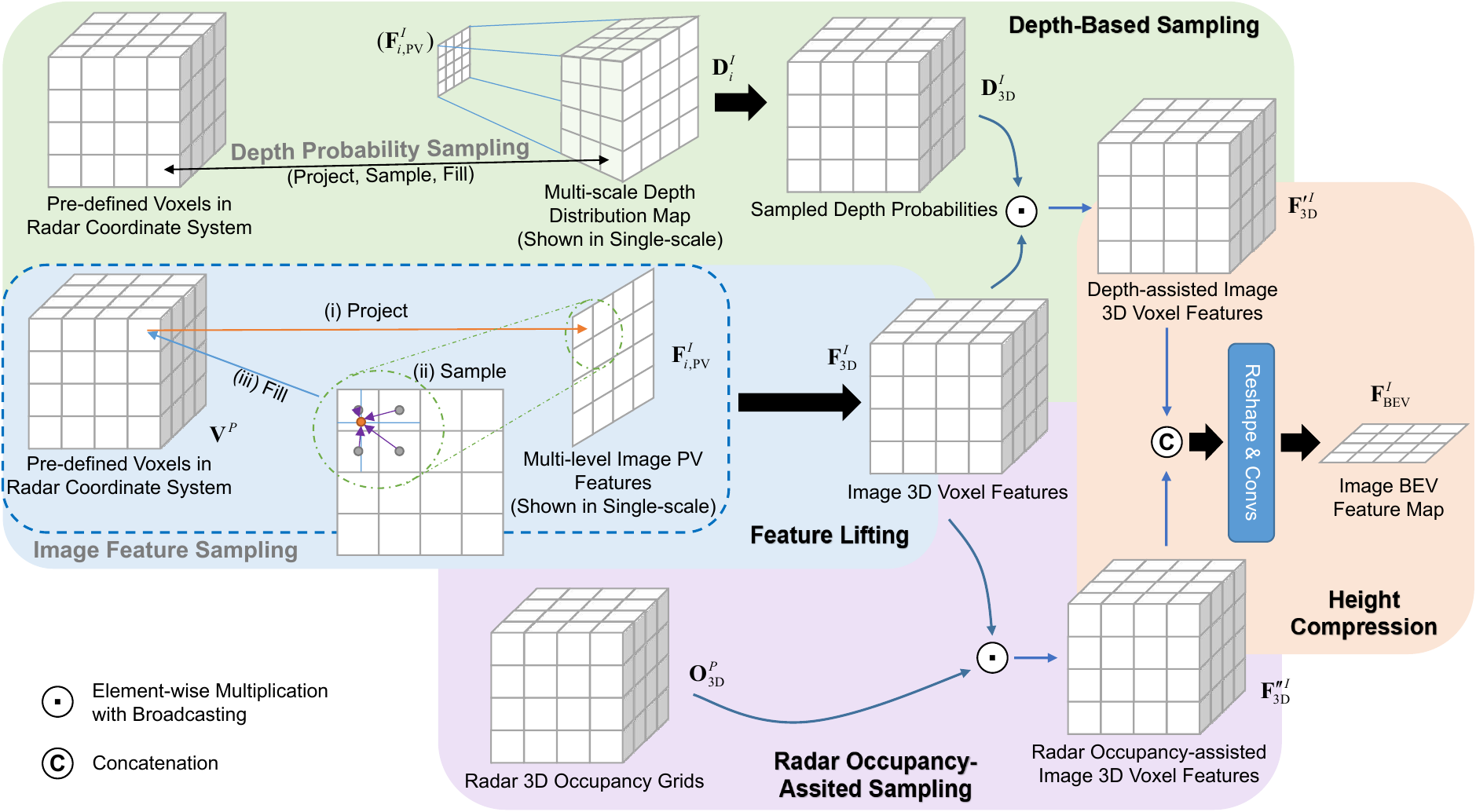}
    \caption{The view transformation module in LXL. During the feature lifting process, image PV features are transformed to 3D voxel features through the ``sampling" strategy. The result is then multiplied by sampled depth probabilities and radar 3D occupancy grids respectively, obtaining depth-assisted image 3D voxel features and radar occupancy-assisted image 3D voxel features. Finally, the two aforementioned image 3D features are concatenated before height compression.}
    \label{view trans}
\end{figure*}

\subsection{View Transformation}
\label{sec view trans}

The process of view transformation, which incorporates predicted multi-scale depth distribution maps and radar 3D occupancy grids, is illustrated in Fig. \ref{view trans}. This method is referred to as ``\textit{radar occupancy-assisted depth-based sampling}" in this paper. To the best of the authors' knowledge, this is the first work to enhance the ``sampling" feature lifting strategy with both predicted image depth distribution maps and radar 3D occupancy grids.

\textbf{Feature Lifting:}
There are two primary strategies for geometrically lifting image features into 3D space. 
The first strategy is ``sampling", where pre-defined 3D voxels are projected onto the image plane, and the features of nearby pixels in the projected region are combined to form the voxel feature. 
Representative models utilizing this strategy include \cite{M2BEV}\cite{Simple-BEV}\cite{OFT}.
The second strategy, ``splatting", involves transforming each image pixel into points or frustum voxels along a straight line in 3D space based on the calibration matrix. The features of these points or frustum voxels are determined by their corresponding pixel features. 
Subsequently, the points are voxelized, or the frustum voxels are transformed into cubic voxels. Prominent models that employ the ``splatting" technique for view transformation include \cite{CaDDN}\cite{BEVDet}\cite{LSS}.

Simple-BEV \cite{Simple-BEV} has demonstrated that the ``sampling" approach outperforms ``splatting", and our experiments in Section \ref{ablation} corroborate this conclusion. 
Therefore, we have chosen the ``sampling" strategy for view transformation in our model. 
Specifically, given the 3-dimensional coordinates of the pre-defined 3D voxels, denoted as $\mathbf{V}^P \in\mathbb{R}^{X\times Y\times Z\times3}$ in the radar coordinate system, the radar-to-image coordinate transformation matrix $\mathbf{T}_\text{r2c}\in\mathbb{R}^{3\times3}$, and the camera intrinsic matrix $\mathbf{I}\in\mathbb{R}^{3\times4}$, we first project the voxel centers onto the image plane using the following equation:
\begin{equation}
    \mathbf{V}^I_{i,j,k}=\mathbf{I}\cdot\mathbf{\bar{T}}_\text{r2c}\cdot\mathbf{\bar{V}}^P_{i,j,k},
\end{equation}
where $\mathbf{\bar{V}}^P_{i,j,k}=[\mathbf{V}^P_{i,j,k},1]\in\mathbb{R}^4$ is the extended coordinates, and 
\begin{equation}
    \mathbf{\bar{T}}_\text{r2c}=\begin{bmatrix}
\mathbf{T}_\text{r2c} & \mathbf{0}\\ 
\mathbf{0} & 1
\end{bmatrix}\in\mathbb{R}^{4\times4}
\end{equation}
is the extended coordinate transformation matrix. $\mathbf{V}^I_{i,j,k}=[ud,vd,d]$ is the projected coordinate in the image coordinate system, where $(u,v)$ and $d$ denotes the pixel index and the image depth, respectively.

Subsequently, the feature of each pre-defined voxel can be obtained through bi-linear sampling on each multi-level image PV feature map. Specifically, we select the $2\times2$ pixels closest to $(u,v)$ and compute the weighted sum of their features, which is then assigned to the corresponding voxel as its feature.
This step is accomplished using the ``$\mathtt{torch.nn.functional.grid\_sample}$" operation, resulting in image 3D voxel features $\mathbf{F}^I_\text{3D}\in\mathbb{R}^{N_\text{lvl}\times X\times Y\times Z\times C_I}$. 

\textbf{Depth-Based Sampling:} However, the aforementioned operations do not consider the predicted image depths, which may lead to sub-optimal feature lifting. While LSS \cite{LSS} employs the outer product for ``depth-based splatting", this approach is not directly applicable in our ``sampling" case due to the different coordinate systems of the predicted depth distribution maps and image 3D features.
To address this issue, we leverage tri-linear sampling, the 3D extension of bi-linear sampling, on the predicted multi-scale image depth distribution maps in the image coordinate system, denoted as $\mathbf{D}_i^I$, to obtain depth probabilities $\mathbf{D}^I_\text{3D}\in\mathbb{R}^{N_\text{lvl}\times X\times Y\times Z}$ for the pre-defined voxels in the radar coordinate system. Concretely, the pre-defined voxels in the radar coordinate system are projected onto the image plane to determine the nearest $2\times2\times2$ image frustum voxels. As each image depth probability corresponds to a frustum voxel, the weighted sum of depth probabilities from the selected frustum voxels is calculated as the sampled depth probability of the pre-defined voxel.

Subsequently, the image 3D voxel features are multiplied by the sampled depth probabilities using the following equation:
\begin{equation}
    \mathbf{F'}^I_\text{3D}=\mathbf{F}^I_\text{3D}\odot\mathbf{D}^I_\text{3D}.
\end{equation}
Here, $\odot$  represents element-wise multiplication with broadcasting, and $\mathbf{F'}^I_\text{3D}\in\mathbb{R}^{N_\text{lvl}\times X\times Y\times Z\times C_I}$ denotes the result of the ``depth-assisted image feature lifting" process.

\textbf{Radar Occupancy-Assisted Sampling:} The model faces challenges in learning accurate depth prediction without direct supervision, as the image depth information is often ambiguous \cite{BEVDepth}. 
To address this issue, one possible approach is to ``generate" depth supervision using radar points. This method involves projecting radar points onto images and assigning their depths as the ground-truth depths of the nearest pixels. However, due to the sparsity of radar point cloud, only a few pixels have ground-truth depth information, and the accuracy of the ground-truth depths are limited because of the noise inherent in radar measurements.

As a result, we choose another approach, leveraging the radar modality differently by adding a branch for lifting the image PV features and fusing with the aforementioned lifted features $\mathbf{F'}^I_\text{3D}$. The latest work of this approach, CRN \cite{CRN}, projects the 2D radar points onto the image plane and applies convolutional operations after pillarization. The resulting convolution output, referred to as the radar occupancy map, is in the image coordinate system, and aids in the view transformation process. 
However, the coordinate transformation and pillarization procedures are time-consuming. In addition, when combing CRN with our ``sampling" strategy, the radar occupancy map must be re-sampled to the radar coordinate system, which further increases the complexity. Thus, our proposed method generates radar occupancy grids in the radar coordinate system directly, as explained in Section \ref{radar branch}. 

It is worth noting that in our model, radar 3D occupancy grids are predicted instead of radar 2D occupancy maps, as 4D radar is capable of capturing height information. Moreover, since the required occupancy grids and radar BEV features share the same BEV resolution, they are generated from the radar BEV features for simplicity.

Specifically, in this ``radar-occupancy assisted sampling" step, the predicted radar 3D occupancy grids $\mathbf{O}^P_\text{3D}$ are multiplied by $\mathbf{F}^I_\text{3D}$ to obtain the radar-assisted image 3D features $\mathbf{F''}^I_\text{3D}\in\mathbb{R}^{N_\text{lvl}\times X\times Y\times Z\times C_I}$ using the following equation:
\begin{equation}
    \mathbf{F''}^I_\text{3D}=\mathbf{F}^I_\text{3D}\odot\mathbf{O}^P_\text{3D}.
\end{equation}

\textbf{Height Compression:} The resulting radar-assisted image 3D features, denoted as $\mathbf{F''}^I_\text{3D}$, and the depth-assisted image 3D features, denoted as $\mathbf{F'}^I_\text{3D}$ are concatenated along the channel dimension and summed along the level dimension. Subsequently, the tensor is reshaped from $X\times Y\times Z\times 2C_I$ to $X\times Y\times (Z\cdot2C_I)$, enabling the application of convolutional layers to facilitate spatial interaction. The process can be mathematically expressed as
\begin{equation}
    \mathbf{F}^I_\text{BEV}=\mathtt{Convs}(\mathtt{Reshape}(\mathtt{Concat}(\mathbf{F'}^I_\text{3D}, \mathbf{F''}^I_\text{3D}))),
\end{equation}
where $\mathbf{F}^I_\text{BEV}\in\mathbb{R}^{X\times Y\times C_I}$ represents the final image BEV features, which are the output obtained from the view transformation module utilizing the \textit{radar occupancy-assisted depth-based sampling} method.

\subsection{Multi-modal Fusion and Detection Head}
\label{fusion and detection head}
After acquiring the radar and image BEV features, the fusion module integrates their information and produces fused BEV feature maps. In our approach, the radar BEV features, denoted as $\mathbf{F}_\text{BEV}^P$, and the image BEV features, denoted as $\mathbf{F}^I_\text{BEV}$, have the same resolution, allowing for concatenation and fusion through convolutional operations. The resulting fused BEV features are subsequently fed into the detection head to predict 3D bounding boxes. In this work, we adopt the methodology of CenterPoint \cite{CenterPoint} to generate category-wise heatmaps and perform object detection. \textit{It is important to note that our fusion strategy and detection head are not limited to specific methods. For instance, our model can also incorporate attention-based fusion techniques and employ an anchor-based detection head.}

\section{Experiments and Analysis}
\label{experiment}

\subsection{Dataset and Evaluation Metrics}
\label{dataset}
\textbf{Dataset:} In this study, we utilize two datasets, View-of-Delft (VoD) \cite{VoD} and TJ4DRadSet \cite{TJ4DRadSet}, to evaluate the performance of our proposed model. These datasets are designed for autonomous driving and encompass data from various sensors, including LiDAR points, 4D radar points, and camera images.  Each object in the datasets is annotated with its corresponding category, a 3D bounding box, and a tracking ID. Moreover, the datasets provide coordinate transformation matrices between different sensors.

The VoD dataset encompasses three object categories we used in experiments: Car, pedestrian, and cyclist. On the other hand, the TJ4DRadSet includes an additional class, truck. It also presents a more diverse range of driving scenarios than VoD. Notably, it exhibits significant variations in lighting conditions throughout the dataset, as well as different road types such as crossroads and elevated roads. Consequently, the 3D object detection task becomes considerably more challenging when working with the TJ4DRadSet dataset.

For both datasets, we adopt the official data splits provided. Specifically, the VoD dataset comprises 5139 frames for training and 1296 frames for validation. Since the official test server for the VoD dataset is not yet released, evaluations and analyses are performed solely on the validation set. In the case of the TJ4DRadSet, the training set consists of 5717 frames, while the test set encompasses 2040 frames.

\textbf{Evaluation Metrics:} Our proposed model is evaluated using specific metrics for each dataset.

For the VoD dataset, there are two official evaluation metrics: AP in the entire annotated area (EAA AP) and AP in the driving corridor (RoI AP). The driving corridor, considered as a region of interest (RoI), is located close to the ego-vehicle and is defined as a specific area, $D_\text{RoI}=\{(x,y,z)|-4\text{m}<x<4\text{m},z<25\text{m}\}$, within the camera coordinate system.  The Intersection over Union (IoU) thresholds used in the calculation of AP are $0.5$, $0.25$, and $0.25$ for cars, pedestrians, and cyclists, respectively. For each predicted bounding box defined as a True Positive (TP), these thresholds determine the minimum overlap required between it and the ground truth.

In the case of the TJ4DRadSet dataset, evaluation metrics include 3D AP and BEV AP for different object classes within a range of $70$ meters. The IoU thresholds for cars, pedestrians, and cyclists follow the same values as those used in the VoD dataset. Additionally, for the truck class, the IoU threshold is set to $0.5$.

\begin{table*}
\centering
\caption{Comparison with state-of-the-art methods on the validation set of VoD \cite{VoD}, where R denotes 4D imaging radar and C indicates camera. The results of methods marked with \dag ~are inherited from \cite{RCFusion}. Note that BEVFusion \cite{bevfusion} is implemented according to the radar + camera configuration file on its official github.}\label{sota comparison}
\begin{tabular}{cc|ccc|c|ccc|c|c}
\hline
\multirow{2}{*}{Method} & \multirow{2}{*}{Modality} & \multicolumn{4}{c|}{AP in the Entire Annotated Area (\%)} & \multicolumn{4}{c|}{AP in the Region of Interest (\%)} & \multirow{2}{*}{FPS} \\ \cline{3-10} 
 &  & Car & Pedestrian & Cyclist & mAP & Car & Pedestrian & Cyclist & mAP &  \\ \hline
PointPillars$^\dag$ (CVPR 2019) \cite{PointPillars} & R & 37.06 &  35.04 & 63.44 & 45.18 & 70.15 & 47.22 & 85.07 & 67.48 & N/A \\
RadarPillarNet$^\dag$ (IEEE T-IM 2023)\cite{RCFusion} & R & 39.30 & 35.10 & 63.63 & 46.01 & 71.65 & 42.80 & 83.14 & 65.86 & N/A \\
LXL - R & R & 32.75 & 39.65 & 68.13 & 46.84 & 70.26 & 47.34 & 87.93 & 68.51  & 44.7 \\
FUTR3D (CVPR 2023) \cite{futr3d} & R+C & \textbf{46.01}  & 35.11 & 65.98 &  49.03 & \textbf{78.66} & 43.10 & 86.19 & 69.32 & 7.3 \\
BEVFusion (ICRA 2023) \cite{bevfusion} & R+C &  37.85   & 40.96  & 68.95   &  49.25   &  70.21  & 45.86  & \textbf{89.48}  & 68.52  & 7.1 \\
RCFusion$^\dag$ (IEEE T-IM 2023) \cite{RCFusion} & R+C & 41.70 & 38.95 & 68.31 & 49.65 & 71.87 & 47.50 & 88.33 & 69.23  & N/A \\
\hline
LXL (\textbf{Ours}) & R+C & 42.33 & \textbf{49.48} & \textbf{77.12} & \textbf{56.31} & 72.18 & \textbf{58.30} & 88.31 & \textbf{72.93}  & 6.1 \\ \hline
\end{tabular}
\end{table*}

\begin{figure*}
    \centering
    \includegraphics[]{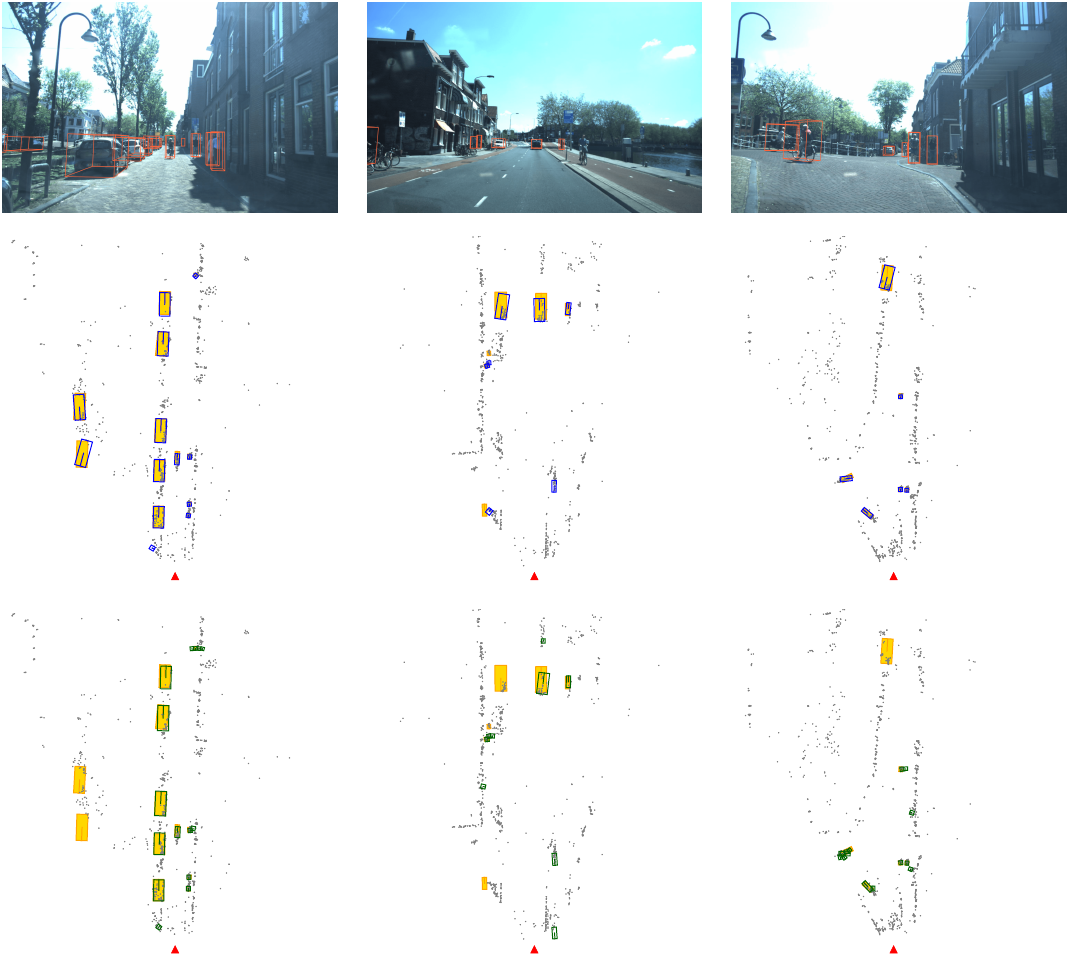}
    \caption{Some visualization results on the VoD \cite{VoD} validation set (best viewed in color and zoom). Each column corresponds to a frame of data containing an image and radar points (gray points) in BEV, where the red triangle denotes the position of the ego-vehicle, and orange boxes represent ground-truths. Blue boxes in the second row stand for predicted bounding boxes from LXL. Note that in the third row, the detection results of LXL-R are also shown as green boxes for comparison.}
    \label{visualization}
\end{figure*}

\subsection{Implementation Details}
\label{implemetation}

The model implementation is based on MMDetection3D \cite{mmdet3d}, an open-source framework for 3D object detection tasks.

\textbf{Hyper-parameter Settings:} The hyper-parameters are determined following the official guidelines of the VoD dataset. The point cloud range (PCR) is set to a specific range, $D_\text{PCR}=\{(x,y,z)|0<x<51.2\text{m}, -25.6\text{m}<y<25.6\text{m}, -3\text{m}<z<2\text{m}\}$, in the radar coordinate system. The pillar size in the voxelization process of radar points is defined as $0.16\text{m}\times0.16\text{m}$. The stride of the radar feature extractor, which consists of the backbone and neck, is adjusted to $2$ to achieve a final BEV resolution of $160\times160$.

For the detection head, we utilize the CenterPoint \cite{CenterPoint} framework. During training, the minimum Gaussian radius for generating ground-truth heatmaps is set to 2. During inference, the top 1000 detections are considered, and a post-processing step with non-maximum suppression (NMS) is applied. The distance thresholds for NMS are set to $4$m, $0.3$m, and $0.85$m for cars, pedestrians, and cyclists, respectively.

Regarding the TJ4DRadSet dataset, the PCR is set to $D_\text{PCR}=\{(x,y,z)|0<x<69.12\text{m}, -39.68\text{m}<y<39.68\text{m}, -4\text{m}<z<2\text{m}\}$, and the other hyper-parameters remain consistent with those used in the VoD dataset. For the truck category, the distance threshold for NMS is $12$m.

\textbf{Training Details:} During training, both images and radar points are normalized with the mean and standard deviation values of the corresponding data in the whole training set before being fed into the model: 
\begin{equation}
    \begin{aligned}
    i & = (i - \mu_I) / \sigma_I,\\
    p_m & = (p_m - (\mu_P)_m) / (\sigma_P)_m,  m \notin \{x, y, z, t\},
    \end{aligned}
\end{equation}
where $i$ is the RGB value of an image pixel, and $p$ is the original feature of a radar point; $I$ and $P$ represent the whole set of image pixels and radar points, and the subscript $m$ denotes the corresponding feature vector component. $\mu_X$ and $\sigma_X$ is the mean and standard deviation of set $X$ ($X=I$ or $P$), respectively. In addition, radar points and ground-truth bounding boxes outside the image view are filtered out to ensure data consistency. This is achieved by projecting radar points and centers of bounding boxes onto the image plane and discarding those falling outside the image. Moreover, random horizontal flipping is applied as a data augmentation technique for both input data and BEV features. 

The model is trained for 80 epochs using the AdamW optimizer and StepLR scheduler. The batch size is set to 6, and the initial learning rate is set to 1e-3. It is important to note that the image backbone and neck are loaded from a pre-trained model, and their parameters are frozen to prevent overfitting. The pre-trained model is from MMDetection \cite{mmdet}, which is YOLOX \cite{YOLOX} trained on the COCO \cite{COCO} dataset for the 2D object detection task.

\subsection{Results and Analysis}
\label{results}
\textbf{Results on VoD:} The experimental results on the VoD \cite{VoD} validation set are presented in Table \ref{sota comparison}. 
We also present the detection accuracy of LXL-R, which is our single-modal baseline without the image branch, occupancy net, and fusion module.
The RoI AP for cars and cyclists is relatively high for LXL-R, indicating that 4D radar alone is effective in perceiving the environment at close range.
However, the RoI AP for pedestrians is limited due to two main reasons. Firstly, pedestrians are small in the BEV representation, often occupying only a single grid or even a fraction of it, making it challenging for the network to accurately regress bounding boxes. 
Additionally, millimeter waves have weak reflections on non-metallic objects, resulting in sparse and less accurate measurements from pedestrians. 
Another observation is that the radar-modal-only model performs poorly in terms of EAA AP for all categories, highlighting the challenges of detecting far-away objects due to the sparsity and noise in radar points.

Upon fusing camera images with radar data, the detection results of different models are improved, particularly in the EAA metric and on pedestrians and cyclists. 
These improvements suggest that dense images with rich semantic information can compensate for the sparsity and noise in radar points, enhancing radar perception for objects that are porous, non-metallic, or located at a distance. 
Compared to RCFusion \cite{RCFusion}, the latest benchmark on 3D object detection with 4D imaging radar and camera fusion, our LXL model achieves higher detection accuracy across almost all categories and evaluation regions.
This indicates that the proposed ``radar occupancy-assisted depth-based samping" strategy is able to achieve precise image view transformation and amplifies the effectiveness of fusion with images.
As our LXL utilize a center-based detection head that is skilled in detecting small objects \cite{3d_object_det_survey_1} while RCFusion adopts an anchor-based head, the performance improvement is higher on pedestrians (+10.5\%) and cyclists (+8.8\%) than that on cars (+0.6\%). 
FUTR3D \cite{futr3d} and BEVFusion \cite{bevfusion} are also implemented on VoD according to the code on their official github, getting 7\% lower results than LXL as these methods are not specially designed for 3D points from 4D radars. 
It is interesting to notice that FUTR3D outperforms LXL in the car category. This can be attributed to the sampling strategy in its transformer decoder head that samples features from the $K$ nearest radar points. However, when the object size is small, there are often not enough points to be sampled, and the sampled features will be polluted by other irrelevant sampled points. Thus, the detection accuracy of pedestrians and cyclists is limited.

The inference speed of LXL and some other models are also listed in Table \ref{sota comparison}, measured on a single V100 GPU. The radar modal-only model, LXL-R, can detect objects in real time, while multi-modal methods all require more than 100ms (FPS $<10$) to generate predictions due to the introduction of images. Although LXL runs slightly slower than FUTR3D \cite{futr3d} and BEVFusion \cite{bevfusion}, the performance gain of more than 7\% makes it relatively acceptable.

Fig. \ref{visualization} showcases the visualization results of our LXL model, demonstrating its accurate detection of various object classes. The predictions of LXL-R are also provided for comparison.
Notably, in some cases, LXL even detects true objects that are not labeled (e.g., the bottom-right cyclist in the second column of the image). 
Moreover, when the radar point cloud is sparse, LXL has the ability to leverage camera information to detect objects that are missed by LXL-R. It is also able to utilize radar measurements to detect objects occluded in the camera view. Therefore, our model effectively leverages the advantages of both modalities to reduce missed detections and improve detection accuracy.

To compare the performance of LiDAR-based 3D object detection and 4D radar-based 3D object detection, Table \ref{lidar comparison} exhibits the experimental results of LXL and LiDAR-based PointPillars \cite{PointPillars}. As expected, the detection accuracy of PointPillars-LiDAR is 19.1\% higher than that of LXL-R, because LiDAR points are much denser and less noisy than 4D radar points, as mentioned in Section \ref{introduction}. However, when fused 4D radar points with camera images, the performance gap is significantly narrowed to 9.6\%, and LXL even outperforms PointPillars-LiDAR in the cyclist category. This result shows that the mutual compensation of low-cost sensors can make 4D radar-camera fusion-based methods have the potential to replace high-cost LiDARs in some aspect.

\begin{figure*}[t]
    \centering
    \includegraphics[]{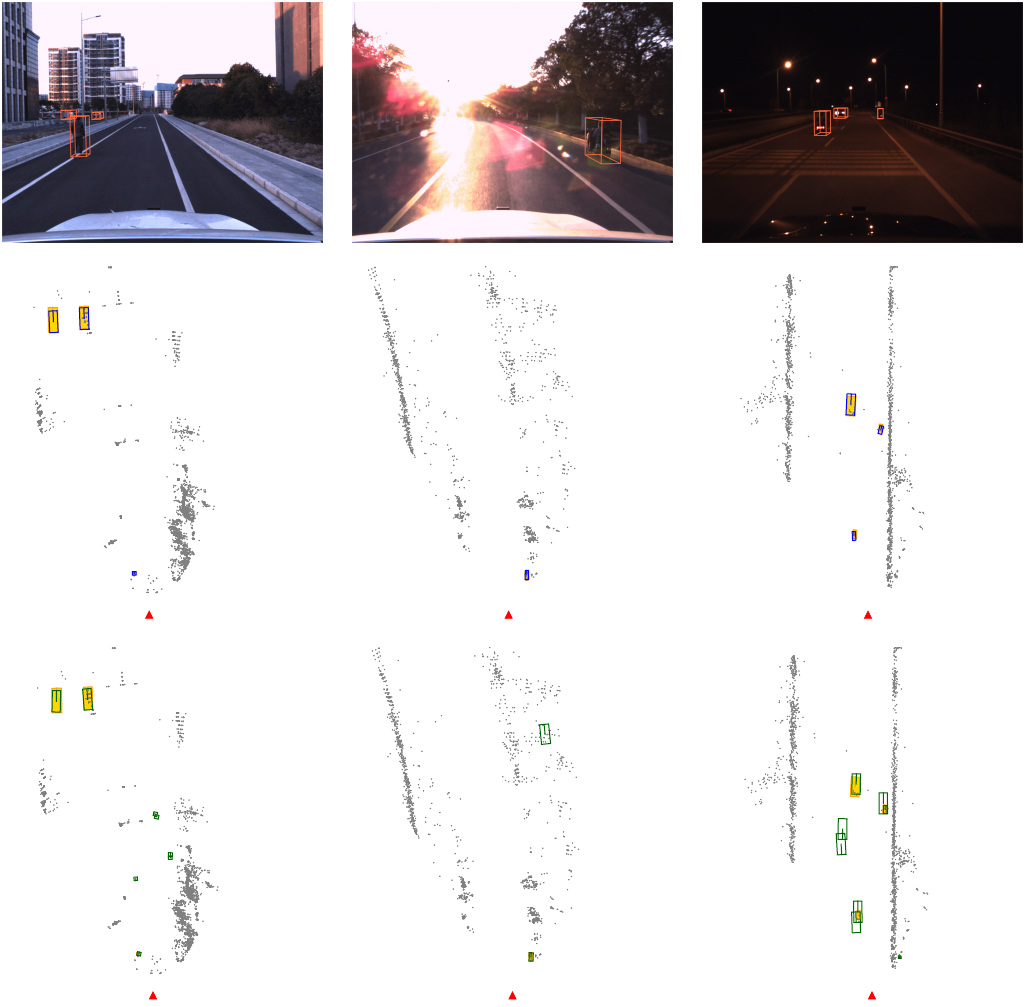}
    \caption{Some visualization results on the TJ4DRadSet \cite{TJ4DRadSet} test set (best viewed in color and zoom). Each column corresponds to a frame of data containing an image and radar points (gray points) in BEV, where the red triangle denotes the position of the ego-vehicle, and orange boxes represent ground-truths. Blue boxes in the second row stand for predicted bounding boxes from LXL. Note that in the third row, the detection results of LXL-R are also shown as green boxes for comparison.}
    \label{visualization_TJ}
\end{figure*}

\textbf{Results on TJ4DRadSet:} To evaluate the generalization ability of our proposed model, we conduct additional experiments on the TJ4DRadSet \cite{TJ4DRadSet} dataset. Table \ref{generalization} presents the performance of different methods on the test set of TJ4DRadSet, where LXL outperforms other 4D radar-camera fusion-based models by more than 2.4\% in 3D mAP. Fig. \ref{visualization_TJ} provides visualizations of the detection results in various scenarios. These results demonstrate the effectiveness of our model in fusing radar and camera information for 3D object detection, even under challenging lighting conditions such as darkness or excessive illumination.

\begin{table}[t]
\fontsize{6.5}{9}\selectfont
\caption{Comparison between classical LiDAR-based PointPillars (the result is inherited from \cite{Reviewing_3D_OD_Approach_for_4D_Radar}) and 4D imaging radar and camera fusion-based LXL on the validation set of VoD \cite{VoD}, where L denotes LiDAR, R denotes 4D imaging radar, C indicates camera.}
\label{lidar comparison}
\centering
\begin{tabular}{cc|cccc}
\hline
\multirow{2}{*}{Method} & \multirow{2}{*}{Modality} & \multicolumn{4}{c}{AP in the Entire Annotated Area (\%)} \\ \cline{3-6}
& & Car & Pedestrian & Cyclist & mAP \\ \hline
LXL - R & R & 32.8 & 39.6 & 68.1 & 46.8 \\
LXL (\textbf{Ours}) & R+C & 42.3 & 49.5 & \textbf{77.1} & 56.3 \\
PointPillars (CVPR 2019) \cite{PointPillars} & L & \textbf{66.6} & \textbf{56.1} & 75.1 & \textbf{65.9} \\
\hline
\end{tabular}
\end{table}

\begin{table}[t]
\scriptsize
\renewcommand{\arraystretch}{1.2}
\caption{Comparison with state-of-the-art methods on the test set of TJ4DRadSet \cite{TJ4DRadSet}, where R denotes 4D imaging radar and C indicates camera. The results of methods marked with \dag ~are inherited from \cite{RCFusion}. Note that BEVFusion \cite{bevfusion} is implemented according to the radar + camera configuration file on its official github.}
\label{generalization}
\centering
\begin{tabular}{cc|cc}
\hline
Method & Modality & 3D mAP (\%) & BEV mAP (\%) \\ \hline
RPFA-Net$^\dag$ (ITSC 2021) \cite{RPFA-Net} & R & 29.91 & 38.94 \\
RadarPillarNet$^\dag$ (T-IM 2023) \cite{RCFusion} & R & 30.37 & 39.24 \\
LXL - R & R & 30.79 & 38.42 \\
FUTR3D (CVPR 2023) \cite{futr3d} & R+C & 32.42 & 37.51 \\
BEVFusion (ICRA 2023) \cite{bevfusion} & R+C & 32.71 & 41.12 \\
RCFusion$^\dag$ (T-IM 2023) \cite{RCFusion} & R+C & 33.85 & 39.76 \\
\hline
LXL (\textbf{Ours}) & R+C & \textbf{36.32} & \textbf{41.20} \\ \hline
\end{tabular}
\end{table}

\begin{table}[t]
\centering
\caption{The performances of our model on different lighting conditions. Note that there are no pedestrians in the ``dark" subset, and the AP is considered as 0 in the computation of the mAP metric.} \label{lighting breakdown}
\begin{tabular}{c|ccc|ccc}
\hline
\multirow{2}{*}{Model} & \multicolumn{3}{c|}{3D mAP (\%)} & \multicolumn{3}{c}{BEV mAP (\%)} \\ \cline{2-7} 
 & Dark & Standard & Shiny & Dark & Standard & Shiny \\ \hline
LXL-R & 20.17 & 30.76 & \textbf{22.68} & 22.36 & 35.47 & \textbf{36.66} \\
LXL & \textbf{22.50} & \textbf{45.74} & 20.33 & \textbf{25.19} & \textbf{49.55} & 28.13 \\ \hline
\end{tabular}
\end{table}

\begin{table}[t]
\centering
\caption{The evaluating results of our model on different distances of objects.} \label{distance breakdown}
\begin{tabular}{c|ccc|ccc}
\hline
\multirow{2}{*}{Model} & \multicolumn{3}{c|}{3D mAP (\%)} & \multicolumn{3}{c}{BEV mAP (\%)} \\ \cline{2-7} 
 & 0-25m & 25-50m & 50-70m & 0-25m & 25-50m & 50-70m \\ \hline
LXL-R & 36.62 & 20.93 & 9.48 & 43.69 & \textbf{28.68} & 14.81 \\
LXL & \textbf{47.30} & \textbf{22.00} & \textbf{14.55} & \textbf{51.69} & 27.08 & \textbf{17.85} \\ \hline
\end{tabular}
\end{table}

To further investigate the influence of lighting conditions and object distances on our LXL model, we analyze the detection results on TJ4DRadSet. 
Specifically, we divide the test set into three subsets based on the brightness of the scenarios: dark, standard, and over-illuminated (referred to as ``Shiny" in Table \ref{lighting breakdown}). These subsets account for approximately 15\%, 60\%, and 25\% of the entire test set, respectively. 
We report the detection accuracy on these subsets in Table \ref{lighting breakdown}. To mitigate the influence of road conditions across subsets, we also include the performance of LXL-R, which is less affected by lighting conditions, in the table.
By comparing the results of LXL with LXL-R on the same subset, we observe that image information is beneficial in normal lighting conditions, as expected.

Interestingly, even in dark scenarios, fusion with images a performance gain of 2.3\% on the 3D mAP because the headlights and taillights of vehicles provide valuable cues for object classification and localization. However, in cases of excessive illumination, the performance deteriorates due to unclear images under such conditions. 
To address this issue, a simple rule-based approach could be employed, such as switching to LXL-R with a single 4D imaging radar mode when the image quality illumination factor falls below a certain threshold. 
As there are few studies about camera and 4D radar fusion-based 3D object detection, we aim to improve the overall performance here, and robustness against image quality degradation is not the primary focus of this work. Improving model robustness will be a subject of our future research.

Furthermore, we evaluate our model on objects at different distances from the ego-vehicle and present the results in Table \ref{distance breakdown}. The LXL model exhibits higher detection accuracy than LXL-R for objects at almost all distances, and the performance decreases as the distance increases due to the sparsity of radar points. Moreover, as the semantic information from images aids in identifying objects at a distance, there are fewer missed detections and more TPs for far-away objects. In contrast, the radar-only modality demonstrates some ability to detect objects at medium range, and the introduction of images primarily improves bounding box regression accuracy. Since the number of TPs significantly impacts the AP, long-range objects benefit more from multi-modal fusion than medium-range objects.

\subsection{Ablation Study}
\label{ablation}

\newcommand{\tabincell}[2]{\begin{tabular}{@{}#1@{}}#2\end{tabular}}
\begin{table*}[]
\centering
\caption{Ablation studies on image feature lifting and radar assistance. Experiments are conducted on VoD \cite{VoD} dataset.
} \label{ablation results}
\begin{threeparttable}[b]
\begin{tabular}{cc|ccc|c|ccc|c}
\hline
 &  & \multicolumn{4}{c|}{AP in the Entire Annotated Area (\%)} & \multicolumn{4}{c}{AP in the Region of Interest (\%)} \\ \cline{3-10} 
\multirow{-2}{*}{\tabincell{c}{Image Feature\\Lifting Method}} & \multirow{-2}{*}{Radar Assistance} & Car & Pedestrian & Cyclist &mAP & Car & Pedestrian & Cyclist &mAP \\ \hline
Splatting & None & 38.05 & 45.01 & 68.12 & 50.39 & 70.73 & 54.92 & 87.28 & 70.98 \\
Sampling & None & \textbf{42.63} & 45.74 & 74.30 & 54.22 & 71.66 & 52.05 & 87.09 & 70.27 \\
Sampling & Depth Supervise & 41.79 & 48.71 & 74.65 & 55.05 & 71.05 & 53.93 & 88.18 & 71.05 \\
Sampling & 3D Occupancy Grids (CRN)\tnote{*} & 42.01 & 45.86 & 76.07 & 54.65 & 71.00 & 50.70 & 87.52 & 69.74 \\
Sampling & 3D Occupancy Grids (\textbf{Ours}) & 42.33 & \textbf{49.48} & \textbf{77.12} & \textbf{56.31} & \textbf{72.18} & \textbf{58.30} & \textbf{88.31} & \textbf{72.93} \\ \hline
\end{tabular}
\begin{tablenotes}
    \item[*] Note: Since additional elevation could be measured by 4D imaging radar but ordinary automotive radar used in CRN \cite{CRN} can only measure range and azimuth of objects, 3D occupancy grids rather than the 2D occupancy maps used in \cite{CRN} are employed.
\end{tablenotes}
\end{threeparttable}
\end{table*}

\begin{figure}
    \centering
    \includegraphics[scale=0.56]{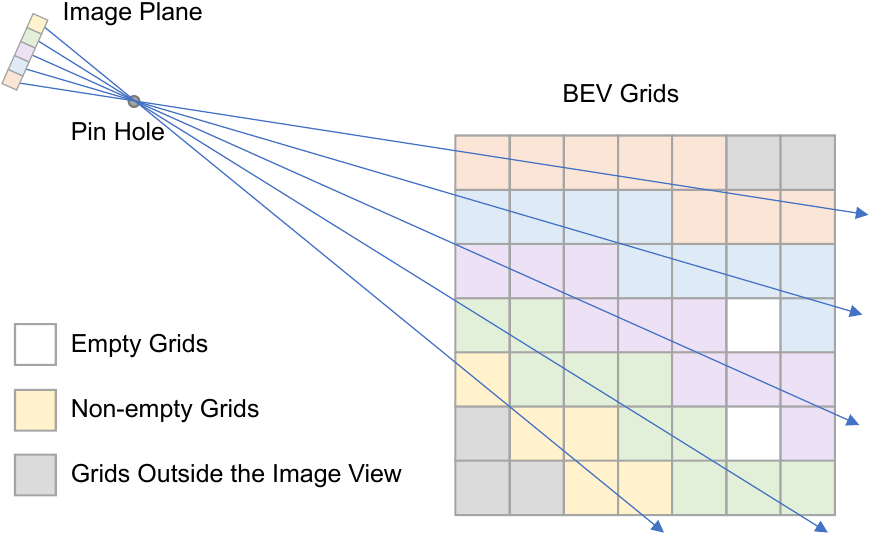}
    \caption{The illustration of \textit{the basic idea} of ``splatting" strategy. For ease of understanding, it is shown in the overhead view. As each image pixel is transformed into radar coordinate system with indefinite depth, it corresponds to a ray in 3D space, named pixel ray. The feature of a BEV grid is related to that of a pixel, if the BEV projection of the pixel ray passes through the grid. Otherwise, the BEV grid is empty. In practice, descretized depth values are predefined and each image pixel only corresponds to several points on the pixel ray, leading to more empty grids in the far distance.}
    \label{splatting}
\end{figure}

In this subsection, we perform several experiments to validate the effectiveness of key design choices in our model on the VoD \cite{VoD} dataset. Specifically, we focus on two aspects: the image feature lifting strategy and the utilization of radar in the image branch. 
We investigate the commonly used geometrical feature lifting strategies, ``sampling" and ``splatting", as described in Section \ref{sec view trans}. For the ``splatting" process, we follow the implementation approach in LSS \cite{LSS}. 
Table \ref{ablation results} presents results of these experiments. While ``sampling" exhibits slightly lower performance in terms of RoI AP compared to ``splatting", it significantly outperforms ``splatting" in terms of EAA APs.  It suggests that while ``splatting" may have a slight advantage at short distances, its performance deteriorates significantly as the distance increases, leading to lower performance compared to ``sampling" in a broader range.
This observation can be attributed to sparsity of splatted point clouds with increasing distance. After pillarization, a considerable number of far-away BEV grids may remain empty, as illustrated in Fig. \ref{splatting}. In contrast, ``sampling" ensures that each 3D voxel is associated with a sampled image feature, as long as the corresponding grid falls within the camera view. Consequently, ``sampling" proves to be more effective in capturing information across a wide range.

Regarding radar assistance in the image branch, we compare our ``radar occupancy-assisted sampling" method with two alternative approaches. One alternative, referred to as ``Depth Supervise" in Table  \ref{ablation results}, is similar to the approach used in BEVDepth \cite{BEVDepth}. It leverages radar points to generate supervision signals for image depth distribution prediction. Specifically, the radar points are first transformed into the image coordinate system. Subsequently, for each projected radar point, we identify the nearest pixel and assign the radar depth as the ground-truth depth for that pixel.
In cases where multiple radar points correspond to a single pixel, we compute the average depth to determine its ground-truth value.
However, we find that reducing the depth loss during training using this method is challenging. This difficulty arises due to the inherent noisiness and sparsity of radar points.
The noise in the radar measurements leads to inaccuracies in the derived ground-truth depths, while the sparsity of radar points poses challenges for the convergence of the depth estimation network.
Consequently, this alternative method only yields a slight improvement in detection accuracy compared to the approach without radar assistance.

An alternative approach involves generating radar 3D occupancy grids by following CRN \cite{CRN}, namely ``3D Occupancy Grids (CRN)" in Table \ref{ablation results}, which differs from our method. For ``3D Occupancy Grids (CRN)", the radar points are projected onto the image plane and voxelized to match the shape of the image depth distribution maps. Subsequently, sparse convolutions are employed to generate radar 3D occupancy grids in the image coordinate system, and tri-linear sampling is applied for image-to-radar coordinate transformation. It is important to note two differences between the aforementioned method and the original approach in CRN \cite{CRN}. Firstly, since the radar points come from 4D radars and contain height information, 3D occupancy grids are generated instead of 2D occupancy maps. Secondly, the feature lifting method here is ``sampling" rather than ``splatting", so the occupancy grids need to be resampled to transform back to the radar coordinate system before being multiplied with the 3D image features. Nonetheless, the underlying idea is the same as CRN.

Compared to the ``Depth Supervise" approach discussed earlier, the performance of the ``3D Occupancy Grids (CRN)" method is even more significantly affected by the sparsity of radar points, due to the larger number of empty grids in 3D space. Furthermore, this method requires a time-consuming projection and voxelization process, whereas our approach only relies on a simple occupancy net to predict radar 3D occupancy grids in the radar coordinate system directly. Consequently, our ``radar occupancy-assisted sampling" strategy offers performance and inference speed advantages.

\section{Conclusion}
\label{conclusion}
In this paper, a new camera and 4D imaging radar fusion model, namely LXL, is proposed for 3D object detection. It is shown that LXL outperforms existing works by a large margin, mainly because its elaborate ``radar occupancy-assisted depth-based sampling" view transformation strategy can effectively transform image PV features into BEV with the aid of predicted image depth distribution maps and radar 3D occupancy grids. This design demonstrates that there is a large room for improving the ``sampling" strategy, where a small enhancement can boost the view transformation significantly.

The proposed LXL provides a framework capable of inspiring subsequent researches in camera and 4D imaging radar fusion-based 3D object detection.
%The future work will focus on strengthening the robustness of LXL by applying an attention-based transformer to achieve adaptive interaction between the two modals.
Future works includes improving the robustness of LXL via attention-based transformer to achieve adaptive interaction between modals, and its applications in subsequent planning and control tasks \cite{wang2023adaptive}\cite{zheng2023control}.

\small
\bibliographystyle{IEEEtran}

\begin{thebibliography}{60}
\bibitem{RadarInsSeg} J. Liu, W. Xiong, L. Bai, Y. Xia, T. Huang, W. Ouyang, and B. Zhu, ``Deep instance segmentation with automotive radar detection points,'' \emph{IEEE Transactions on Intelligent Vehicles}, vol. 8, no. 1, pp. 84-94, 2023. 

\bibitem{RadarInsSeg2} W. Xiong, J. Liu, Y. Xia, T. Huang, B. Zhu, and W. Xiang, ``Contrastive learning for automotive mmWave radar detection points based instance segmentation,'' in \emph{Proceedings of the IEEE International Conference on Intelligent Transportation Systems (ITSC)}, 2022, pp. 1255-1261. 

\bibitem{RaLiBEV} Y. Yang, J. Liu, T. Huang, Q.-L. Han, G. Ma, and B. Zhu, ``RaLiBEV: Radar and LiDAR BEV Fusion Learning for Anchor Box Free Object Detection Systems," 2022, \emph{arXiv:2211.06108.}

\bibitem{GNN-PMB} 
J. Liu, L. Bai, Y. Xia, T. Huang, B. Zhu, and Q.-L. Han, ``GNN-PMB: A simple but effective online 3D multi-object tracker without bells and whistles," \emph{IEEE Transactions on Intelligent Vehicles}, vol. 8, no. 2, pp. 1176-1189, 2023. 

\bibitem{Sun2023vision}
P. Sun, S. Li, B. Zhu, Z. Zuo, X. Xia, ``Vision-Based Fixed-Time Uncooperative Aerial Target Tracking for UAV,'' \emph{IEEE/CAA Journal of Automatica Sinica}, vol. 10, no. 5, pp. 1322-1324, 2023. 

\bibitem{3d_object_det_survey_1}
J. Mao, S. Shi, X. Wang, and H. Li, ``3D object detection for autonomous driving: A review and new outlooks," \emph{International Journal of Computer Vision (IJCV)}, vol. 131, pp. 1909–1963, 2023.

\bibitem{automotive_radar_survey}
S. Sun, A. P. Petropulu, and H. V. Poor, ``MIMO radar for advanced driver-assistance systems and autonomous driving: Advantages and challenges,'' \emph{IEEE Signal Processing Magazine}, vol. 37, no. 4, pp. 98–117, 2020. 

\bibitem{cooperative_perception_survey_tits}
A. Caillot, S. Ouerghi, P. Vasseur, R. Boutteau, and Y. Dupuis, ``Survey on cooperative perception in an automotive context,'' \emph{IEEE Transactions on Intelligent Transportation Systems}, vol. 23, no. 9, pp. 14204-14223, 2022.

\bibitem{4D_radar_overview}
Z. Han, J. Wang, Z. Xu, S. Yang, L. He, S. Xu, and J. Wang, ``4D millimeter-wave radar in autonomous driving: A survey,'' 2023, \emph{arXiv:2306.04242.}

\bibitem{VoD}
A. Palffy, E. Pool, S. Baratam, J. F. Kooij, and D. M. Gavrila, ``Multi-class road user detection with 3+1D radar in the View-of-Delft dataset,'' \emph{IEEE Robotics and Automation Letters}, vol. 7, no. 2, pp. 4961–4968, 2022.

\bibitem{RPFA-Net}
B. Xu, X. Zhang, L. Wang, X. Hu, Z. Li, S. Pan, J. Li, and Y. Deng, ``RPFA-Net: A 4D radar pillar feature attention network for 3D object detection,'' in \emph{Proceedings of the IEEE International Intelligent Transportation Systems Conference (ITSC)}, 2021, pp. 3061–3066. 

\bibitem{RadarMFNet}
B. Tan, Z. Ma, X. Zhu, S. Li, L. Zheng, S. Chen, L. Huang, and J. Bai, ``3D object detection for multi-frame 4D automotive millimeter-wave radar point cloud,'' \emph{IEEE Sensors Journal}, 2022, doi: 10.1109/JSEN.2022.3219643. 

\bibitem{3d_object_det_survey_2}
M. Drobnitzky, J. Friederich, B. Egger, and P. Zschech, ``Survey and systematization of 3D object detection models and methods," \emph{The Visual Computer}, pp. 1-47, 2023.

\bibitem{3d_object_det_survey_3}
Y. Ma, T. Wang, X. Bai, H. Yang, Y. Hou, Y. Wang, Y. Qiao, R. Yang, D. Manocha, and X. Zhu, ``Vision-centric BEV perception: A survey," 2022, \emph{arXiv:2208.02797.}

\bibitem{CaDDN}
C. Reading, A. Harakeh, J. Chae, and S. L. Waslander, ``Categorical depth distribution network for monocular 3D object detection,'' in \emph{Proceedings of the IEEE/CVF Conference on Computer Vision and Pattern Recognition (CVPR)}, 2021, pp. 8555–8564.

\bibitem{BEVDet}
J. Huang, G. Huang, Z. Zhu, and D. Du, ``BEVDet: High-performance multi-camera 3D object detection in bird-eye-view,'' 2023, \emph{arXiv:2112.11790.}

\bibitem{M2BEV}
E. Xie, Z. Yu, D. Zhou, J. Philion, A. Anandkumar, S. Fidler, P. Luo, and J. M. Alvarez, ``M2BEV: Multi-camera joint 3D detection and segmentation with unified birds-eye view representation,'' 2022, \emph{arXiv:2204.05088.}

\bibitem{BEVFormer}
Z. Li, W. Wang, H. Li, E. Xie, C. Sima, T. Lu, Y. Qiao, and J. Dai, ``BEVFormer: Learning bird’s-eye-view representation from multi-camera images via spatio-temporal transformers,'' in \emph{Proceedings of the 17th European Conference on Computer Vision (ECCV)}. Springer, 2022, pp. 1–18.

\bibitem{PolarFormer}
Y. Jiang, L. Zhang, Z. Miao, X. Zhu, J. Gao, W. Hu, and Y.-G. Jiang, ``PolarFormer: Multi-camera 3D object detection with polar transformers,'' 2022, \emph{arXiv:2206.15398.}

\bibitem{LSS}
J. Philion and S. Fidler, ``Lift, Splat, Shoot: Encoding images from arbitrary camera rigs by implicitly unprojecting to 3D,'' in \emph{Proceedings of the 16th European Conference on Computer Vision (ECCV)}. Springer, 2020, pp. 194–210.

\bibitem{BEVDepth}
Y. Li, Z. Ge, G. Yu, J. Yang, Z. Wang, Y. Shi, J. Sun, and Z. Li, ``BEVDepth: Acquisition of reliable depth for multi-view 3D object detection,'' 2022, \emph{arXiv:2206.10092.}

\bibitem{CRN}
Y. Kim, S. Kim, J. Shin, J. W. Choi, and D. Kum, ``CRN: Camera radar net for accurate, robust, efficient 3D perception,'' in 
\emph{Proceedings of the IEEE/CVF International Conference on Computer Vision (ICCV)}, 2023.
%\emph{Proceedings of the International Conference on Learning Representations (ICLR), Workshop on Scene Representations for Autonomous Driving}, 2023.

\bibitem{Simple-BEV}
A. W. Harley, Z. Fang, J. Li, R. Ambrus, and K. Fragkiadaki, ``Simple-BEV: What really matters for multi-sensor BEV perception?'' in \emph{Proceedings of the IEEE International Conference on Robotics and Automation (ICRA)}, 2023.

\bibitem{TJ4DRadSet}
L. Zheng, Z. Ma, X. Zhu, B. Tan, S. Li, K. Long, W. Sun, S. Chen, L. Zhang, M. Wan, et al., ``TJ4DRadSet: A 4D radar dataset for autonomous driving,'' in \emph{Proceedings of the IEEE 25th International Conference on Intelligent Transportation Systems (ITSC)}. IEEE, 2022, pp. 493–498.

\bibitem{M3D-RPN}
G. Brazil and X. Liu, ``M3D-RPN: Monocular 3D region proposal network for object detection,'' in \emph{Proceedings of the IEEE/CVF International Conference on Computer Vision (ICCV)}, 2019, pp. 9287–9296.

\bibitem{IPM}
H. A. Mallot, H. H. B\"{u}lthoff, J. Little, and S. Bohrer, ``Inverse perspective mapping simplifies optical flow computation and obstacle detection,'' \emph{Biological Cybernetics}, vol. 64, no. 3, pp. 177–185, 1991.

\bibitem{Pseudo-LiDAR}
Y. Wang, W.-L. Chao, D. Garg, B. Hariharan, M. Campbell, and K. Q. Weinberger, ``Pseudo-LiDAR from visual depth estimation: Bridging the gap in 3D object detection for autonomous driving,'' in \emph{Proceedings of the IEEE/CVF Conference on Computer Vision and Pattern Recognition (CVPR)}, 2019, pp. 8445–8453.

\bibitem{Cam3DRadFusion1}
T.-Y. Lim, A. Ansari, B. Major, D. Fontijne, M. Hamilton, R. Gowaikar, and S. Subramanian, ``Radar and camera early fusion for vehicle detection in advanced driver assistance systems,'' in \emph{Proceedings of the 33rd Conference on Neural Information Processing Systems (NeurIPS), Machine Learning for Autonomous Driving Workshop}, vol. 2, 2019, p. 7.

\bibitem{Cam3DRadFusion2}
R. Nabati and H. Qi, ``Radar-camera sensor fusion for joint object detection and distance estimation in autonomous vehicles,'' 2020, \emph{arXiv:2009.08428.}

\bibitem{CenterFusion}
R. Nabati and H. Qi, ``CenterFusion: Center-based radar and camera fusion for 3D object detection,'' in \emph{Proceedings of the IEEE/CVF Winter Conference on Applications of Computer Vision (WACV)}, 2021, pp. 1527–1536.

\bibitem{radar_camera_od_seg_survey}
S. Yao, R. Guan, X. Huang, Z. Li, X. Sha, Y. Yue, E. G. Lim, H. Seo, K. L. Man, X. Zhu, and Y. Yue, ``Radar-camera fusion for object detection and semantic segmentation in autonomous driving: A comprehensive review,'' \emph{IEEE Transactions on Intelligent Vehicles}, 2023. doi: 10.1109/TIV.2023.3307157.

\bibitem{CramNet}
J.-J. Hwang, H. Kretzschmar, J. Manela, S. Rafferty, N. Armstrong-Crews, T. Chen, and D. Anguelov, ``CramNet: Camera-radar fusion with ray-constrained cross-attention for robust 3D object detection,'' in \emph{Proceedings of the 17th European Conference on Computer Vision (ECCV)}. Springer, 2022, pp. 388–405.

\bibitem{RCBEV}
T. Zhou, J. Chen, Y. Shi, K. Jiang, M. Yang, and D. Yang, ``Bridging the view disparity between radar and camera features for multi-modal fusion 3D object detection,'' \emph{IEEE Transactions on Intelligent Vehicles}, vol. 8, no. 2, pp. 1523–1535, 2023.

\bibitem{RADIANT}
Y. Long, A. Kumar, D. Morris, X. Liu, M. Castro, and P. Chakravarty, ``RADIANT: Radar-image association network for 3D object detection,'' in \emph{Proceedings of the 37th AAAI Conference on Artificial Intelligence}, 2023.

\bibitem{MVFusion}
Z. Wu, G. Chen, Y. Gan, L. Wang, and J. Pu, ``MVFusion: Multi-view 3D object detection with semantic-aligned radar and camera fusion,'' in \emph{Proceedings of the IEEE International Conference on Robotics and Automation (ICRA)}, 2023, pp. 2766-2773.

\bibitem{CRAFT}
Y. Kim, S. Kim, J. W. Choi, and D. Kum, ``CRAFT: Camera-radar 3D object detection with spatio-contextual fusion transformer,'' 2022, \emph{arXiv:2209.06535. Accepted by the 37th AAAI Conference on Artificial Intelligence.}

\bibitem{TransCAR}
P. Su, M. Daniel, and R. Hayder, ``TransCAR: Transformer-based camera-and-radar fusion for 3D object detection,'' 2023, \emph{arXiv:2305.00397. Accepted by the IEEE/RSJ International Conference on Intelligent Robots and Systems (IROS).}

\bibitem{Astyx}
M. Meyer and G. Kuschk, ``Automotive radar dataset for deep learning based 3D object detection,'' in \emph{Proceedings of the IEEE 16th European Radar Conference (EuRAD)}, 2019, pp. 129–132.

\bibitem{RADIal}
J. Rebut, A. Ouaknine, W. Malik, and P. P\'{e}rez, ``Raw high-definition radar for multi-task learning,'' in \emph{Proceedings of the IEEE/CVF Conference on Computer Vision and Pattern Recognition (CVPR)}, 2022, pp. 17021–17030.

\bibitem{K-radar}
D.-H. Paek, S.-H. Kong, and K. T. Wijaya, ``K-radar: 4D radar object detection for autonomous driving in various weather conditions,'' in \emph{Proceedings of the 36th Conference on Neural Information Processing Systems (NeuIPS), Datasets and Benchmarks Track}, 2022.

\bibitem{SECOND}
Y. Yan, Y. Mao, and B. Li, ``SECOND: Sparsely embedded convolutional detection,'' \emph{Sensors}, vol. 18, no. 10, p. 3337, 2018.

\bibitem{CenterPoint}
T. Yin, X. Zhou, and P. Krahenbuhl, ``Center-based 3D object detection and tracking,'' in \emph{Proceedings of the IEEE/CVF Conference on Computer Vision and Pattern Recognition (CVPR)}, 2021, pp. 11784–11793.

\bibitem{PointPillars}
A. H. Lang, S. Vora, H. Caesar, L. Zhou, J. Yang, and O. Beijbom, ``PointPillars: Fast encoders for object detection from point clouds,'' in \emph{Proceedings of the IEEE/CVF Conference on Computer Vision and Pattern Recognition (CVPR)}, 2019, pp. 12697–12705.

\bibitem{PointNet}
C. R. Qi, H. Su, K. Mo, and L. J. Guibas, ``PointNet: Deep learning on point sets for 3D classification and segmentation,'' in \emph{Proceedings of the IEEE Conference on Computer Vision and Pattern Recognition (CVPR)}, 2017, pp. 652–660.

\bibitem{SMURF}
J. Liu, Q. Zhao, W. Xiong, T. Huang, Q.-L. Han, and B. Zhu, ``SMURF: Spatial multi-representation fusion for 3D object detection with 4D imaging radar,'' 2023, \emph{arXiv:2307.10784.}

\bibitem{InterFusion}
L. Wang, X. Zhang, B. Xv, J. Zhang, R. Fu, X. Wang, L. Zhu, H. Ren, P. Lu, J. Li, and H. Liu, ``InterFusion: Interaction-based 4D radar and LiDAR fusion for 3D object detection,'' in \emph{Proceedings of the IEEE/RSJ International Conference on Intelligent Robots and Systems (IROS)}, 2022, pp. 12247–12253. 

\bibitem{M2-Fusion}
L. Wang, X. Zhang, J. Li, B. Xv, R. Fu, H. Chen, L. Yang, D. Jin, and L. Zhao, ``Multi-modal and multi-scale fusion 3D object detection of 4D radar and LiDAR for autonomous driving,'' \emph{IEEE Transactions on Vehicular Technology}, pp. 1–15, 2022.

\bibitem{self-attentions}
A. Vaswani, N. Shazeer, N. Parmar, J. Uszkoreit, L. Jones, A. N. Gomez,  and I. Polosukhin, ``Attention is all you need,''  \emph{Advances in Neural Information Processing Systems}, 2017, pp. 30.

\bibitem{Cam4DRadFusion}
H. Cui, J. Wu, J. Zhang, G. Chowdhary, and W. R. Norris, ``3D detection and tracking for on-road vehicles with a monovision camera and dual low-cost 4D mmwave radars,'' in \emph{Proceedings of the IEEE International Intelligent Transportation Systems Conference (ITSC)}. IEEE, 2021, pp. 2931–2937.

\bibitem{SSMA}
A. Valada, R. Mohan, and W. Burgard, ``Self-supervised model adaptation for multi-modal semantic segmentation,'' \emph{International Journal of Computer Vision}, vol. 128, no. 5, pp. 1239–1285, 2020.

\bibitem{RCFusion}
L. Zheng, S. Li, B. Tan, L. Yang, S. Chen, L. Huang, J. Bai, X. Zhu, and Z. Ma, ``RCFusion: Fusing 4D radar and camera with bird’s-eye view features for 3D object detection,'' \emph{IEEE Transactions on Instrumentation and Measurement}, 2023, doi: 10.1109/TIM.2023.3280525.

\bibitem{YOLOX}
Z. Ge, S. Liu, F. Wang, Z. Li, and J. Sun, ``YOLOX: Exceeding YOLO series in 2021,'' 2021, \emph{arXiv:2107.08430.}

\bibitem{CSPNet}
C.-Y. Wang, H.-Y. M. Liao, Y.-H. Wu, P.-Y. Chen, J.-W. Hsieh, and I.-H. Yeh, ``CSPNet: A new backbone that can enhance learning capability of CNN,'' in \emph{Proceedings of the IEEE/CVF Conference on Computer Vision and Pattern Recognition (CVPR) workshops}, 2020, pp. 390–391.

\bibitem{PAN}
S. Liu, L. Qi, H. Qin, J. Shi, and J. Jia, ``Path aggregation network for instance segmentation,'' in \emph{Proceedings of the IEEE Conference on Computer Vision and Pattern Recognition (CVPR)}, 2018, pp. 8759–8768.

\bibitem{OFT}
T. Roddick, A. Kendall, and R. Cipolla, ``Orthographic feature transform for monocular 3D object detection,'' in \emph{Proceedings of the British Machine Vision Conference (BMVC)}. BMVA Press, September 2019, pp. 59.1–59.13.

\bibitem{mmdet3d}
MMDetection3D Contributors, ``MMDetection3D: OpenMMLab next generation platform for general 3D object detection,'' https://github.com/open-mmlab/mmdetection3d, 2020.


\bibitem{futr3d}
X. Chen, T. Zhang, Y. Wang, Y. Wang, and H. Zhao, ``FUTR3D: A unified sensor fusion framework for 3D detection,'' in \emph{Proceedings of the IEEE/CVF Conference on Computer Vision and Pattern Recognition (CVPR)}, 2023, pp. 172–181.

\bibitem{bevfusion}
Z. Liu, H. Tang, A. Amini, X. Yang, H. Mao, D. Rus, and S. Han,
``BEVFusion: Multi-task multi-sensor fusion with unified bird’s-eye view representation,'' in \emph{Proceedings of the IEEE International Conference on Robotics and Automation (ICRA)}, 2023, pp. 2774-2781.

\bibitem{mmdet}
K. Chen, J. Wang, J. Pang, Y. Cao, Y. Xiong, X. Li, S. Sun, W. Feng, Z. Liu, J, Xu, Z. Zhang, D. Cheng, C. Zhu, T. Cheng, Q. Zhao, B. Li, X. Lu, R. Zhu, Y. Wu, J. Dai, J. Wang, J. Shi, W. Ouyang, C. Loy, and D. Lin ``MMDetection: Open MMLab detection toolbox and benchmark," 2019, \emph{arXiv:1906.07155.}

\bibitem{COCO}
T.-Y. Lin, M, Maire, S. Belongie, J. Hays, P. Perona, D. Ramanan, P. Doll{\'a}r, and C. L. Zitnick, ``Microsoft COCO: Common objects in context," in \emph{Proceedings of the 13th European Conference on Computer Vision (ECCV)}. Springer, 2014, pp. 740-755.

\bibitem{Reviewing_3D_OD_Approach_for_4D_Radar}
P. Palmer, M. Krueger, R. Altendorfer, G. Adam, and T. Bertram, ``Reviewing 3D object detectors in the context of high-resolution 3+1D radar," in \emph{Proceedings of the Conference on Computer Vision and Pattern Recognition (CVPR), Workshop on 3D Vision and Robotics}, 2023. 

\bibitem{wang2023adaptive}
J. Wang, B. Zhu, and Z. Zheng, ``Robust Adaptive Control for a Quadrotor UAV With Uncertain Aerodynamic Parameters,'' \emph{IEEE Transactions on Aerospace and Electronic Systems} (early access), 2023, doi: 10.1109/TAES.2023.3303133.

\bibitem{zheng2023control}
W. Zheng, and B. Zhu, ``Control Lyapunov–Barrier function based model predictive control for stochastic nonlinear affine systems,'' \emph{IEEE International Journal of Robust and Nonlinear Control} (early access), 2023, doi: doi.org/10.1002/rnc.6962.


\end{thebibliography}

\end{document}